\documentclass[10pt]{amsart}
\usepackage{kept}

\begin{document}
\title{Hrebs and Cohesion Chains   as  similar tools  for   semantic text properties research}

\author{ Doina Tatar$^{(1)}$}
\address{$^{1}$ University "Babes-Bolyai", Romania}
\email{dtatar@cs.ubbcluj.ro}

\author{ Mihaiela Lupea$^{(2)}$}
\address{$^{2}$  University "Babes-Bolyai", Romania }
\email{lupea@cs.ubbcluj.ro}

\author{ Epaminondas Kapetanios$^{(3)}$}
\address{$^{3}$ University Westminster, UK}
\email{E.Kapetanios@westminster.ac.uk}


\date{}

\begin{abstract}
  In this study it is proven that the Hrebs used in  Denotation analysis of texts and Cohesion Chains (defined as a fusion between  Lexical Chains  and  Coreference Chains) represent similar  linguistic tools. This result gives us the possibility to extend to
  Cohesion Chains (CCs) some important  indicators as, for example the  Kernel of  CCs,  the topicality of a  CC, text concentration, CC-diffuseness and mean diffuseness of the text. Let us mention that nowhere in the Lexical Chains or Coreference Chains literature these kinds of indicators are introduced and used since now.  Similarly, some applications of CCs in the study of a text (as for example  segmentation or summarization of a text) could be realized starting from hrebs. As an   illustration  of the similarity between Hrebs and CCs   a detailed analyze of the poem "Lacul" by Mihai Eminescu is given.

\end{abstract}


 \maketitle

\section{Introduction}

Denotation analysis is a complex discipline concerned with the mutual relationships of sentences.   An important tool used in Denotation analysis is the concept of {\it hreb}  defined in \cite{Popescu et al. 2013}   as  a discontinuous text unit that can be presented in a set form or a list form, when the order is important. A hreb contains all entities denoting the same real entity or referring to one another in the text.
This basic concept is  baptized in this way in honor of L. Høebíèek (\cite{Hreb}) who
introduced measurement in the domain of Denotation analysis, as it is known in Quantitative Linguistics.
As we will show, the concepts as   Lexical Chain or
 Coreference Chain (as in  Computational Linguistics) subsume the notion of hrebs in the variant of word-hrebs.
 In fact, we are interested  in this paper  only in   the notion of word-hrebs (for other kinds of hrebs:
  morpheme-hrebs,  phrase-hrebs and sentence-hrebs
 see \cite{Popescu et al. 2013}, \cite{Ziegler and Altmann}).


 We will  operate with  the concept of Cohesion Chain (CC), defined as   a Lexical Chain {\bf or} a Coreference Chain, and will show the relationship between CCs and hrebs (more exactly a slow modified  kind of word-hrebs, quasi-hrebs). Due to this relation, some denotational properties of a text defined using hrebs could be translated to CCs, in the benefit of the last ones. Similarly, some applications of CCs in the study of a text (as for example  segmentation or summarization of a text) could be realized starting from quasi-hrebs.

   The structure of the paper is as follows:  Section 2  presents the concept of hreb and some indicators of a text connected with it. In Section 3  the Lexical Chains, the Coreference Chains, and their use in segmentation and summarization are introduced.
In Section 4 we analyze a  poem by Eminescu from the point of view of word-hrebs (as in \cite{Popescu et al. 2013}) and CCs. The paper ends with some conclusions and further work proposal.

\section{Hrebs }

	A word-hreb contains all the words which are synonyms or refer to one of the synonyms.
	The hrebs usually are constructed using some  rules such that
a word  belongs to one or more hrebs \cite{Popescu et al. 2013}. For example a verb with personal ending (1st and 2nd person) belongs both to the given verb and to the person (subject) it overtly refer to.  We will slightly modify the definition of a hreb eliminating the above syntactical constraint  and will denote the new concept by {\it quasi-hreb}.   Namely,
for us  verbs with personal ending (1st and 2nd person) belong  to the given verb  and don't have any connection with the hreb representing  the subject of these verbs. In this way, a word belongs to only one {\it quasi-hreb}, similarly with the property that a word belongs to only one Lexical Chain or Reference Chain (Coherence Chain).
The rest of the properties of hrebs mentioned in  \cite{Popescu et al. 2013} are unmodified  for  {\it quasi-hreb}:
references belong to the quasi-hreb of the word they refer to, e.g. pronouns and Named Entities  belong to the basic word; synonyms constitute a common quasi-hreb;
articles and prepositions are  not considered;
adverbs may coincide with adjectives,  and may belong to the same quasi-hreb.

	
	According to the information and ordering of entities,   \cite{Ziegler and Altmann} defines five kinds of hrebs:

(1) Data-hreb containing the raw data, e.g. words, and the position of each unit in text.

(2) List-hreb containing the data but without the positions of the units in the text.

(3) Set-hreb being the set containing only the lemmas (for word-hrebs).

(4) Ordered set-hreb is identical with (3) but the units are ordered according to a certain principle, e.g. alphabetically, or according to length, frequency, etc.

(5) Ordered position-hreb containing only the positions of units in the given text.

 In our example in Section 4 we will use only the cases 1, 2 and 3.

Complete word-hreb analyses of several texts can be found in \cite{Ziegler and Altmann}.

\subsection{Denotational analysis with hrebs}

Creating hrebs means a reduction of the text to its fundamental semantic components. Having defined them one can make statements both about the text and the hrebs themselves and obtain new indicators.
A short introduction in these indicators is given below (for a complete presentation see
\cite{Popescu et al. 2013}):

1.	By lemmatizing the words occurring in a List-hreb, and eliminating the duplicates,   the corresponding Set-hreb is obtained. If in a Set-hreb there are at least two words (different lemmas), then the hreb belongs to the $Kernel$ (core) of the text, i.e. if $|{hreb_i}| \geq 2 $ then $hreb_i \in    Kernel$.  The hrebs of a $Kernel$ will be called $kernel\,\, hrebs$.

2.	An important indicator of a text is the size of the $Kernel$, denoted by $|Kernel|$.

3.	Topicality of a set-$kernel\,\,  hreb$  $H_i$,   is calculated as:

   $$ T(H_i)=\frac{|H_i|}{|Kernel|}$$

4.   Kernel concentration is defined as the size of the kernel divided by the total number $n$  of hrebs in the text:

     $$KC= \frac{|Kernel|}{n}$$

5. Text concentration is calculated based on the List-hrebs. If $H_i$ is a List-hreb (containing all word-forms, not only lemmas) and $L$ is the number of tokens in the text, then $p_i=|H_i|/L$ is the relative frequency of the List-hreb  $H_i$. Text concentration $TC$ is given as:

    $$TC= \sum_{i=1}^n  p_i ^2 $$

 Relative text concentration, $TC_{rel}$ is defined as:

     $$TC_{rel}= \frac{1-\sqrt{TC}}{1-1/\sqrt{n}}$$

    6. Hreb diffuseness

  The diffuseness $D_H$ of a  given hreb $H$ with $nH$ elements,  where the positions of tokens are (in an ascending  order)  $P=\{pos_1, ...,pos_{nH}\} $,  is defined using the maximal and minimal position of tokens  occurring in it:

      $$D_H=\frac{pos_{nH} - pos_1}{nH} $$

\noindent i.e.  the difference of the last and the first position divided by the cardinal number of the hreb.

	7. Mean diffuseness of the text is:

      $$ D_{Text}= \frac{1}{K} \sum_{j=1}^K D_{H_j} $$

\noindent where $K$ is the number of kernel-hrebs ($|Kernel|$)  in $Text$.

    8.    Finally, text compactness is defined as:

      $$C= \frac{1-n/L}{1-1/L} $$

\noindent where $n$ is the number of hrebs in the text and $L$ is the number of (word-)tokens.

\section{Cohesion Chains}

\subsection{Lexical Chains}

Lexical Chains (LCs) are sequences of words which are in a lexical cohesion relation with each other and they tend to indicate
portions of a text that form  semantic units (\cite{Okumura and Honda 1994}, \cite{Stokes et al. 2004}, \cite{Labadie and Prince 2008}). The most frequent lexical cohesion relations are the synonymy and the repetition, but could be also  hypernyms, hyponyms, etc.. Lexical cohesion relationships between the words of LCs are established using an auxiliary knowledge source such as a dictionary or a thesaurus.

    A Lexical Chain could be formalized as:

       $$ LC_i:[LC_i^1 \,\,(Token_j), \cdots, LC_i^m \,\, (Token_k)] $$

\noindent    where the first element of the chain $LC_i$ is  the word $ LC_i^1$, representing the token with the number $j$ in the text,  the last  element of the chain $LC_i$ is  the word $ LC_i^m  $, representing the token with the number $k$ in the text (where $j < k$), the length of the chain $   LC_i$ is $m$. Because
the analyze is made on the level of sentences,  usually
    the sentences where the words occur are indicated. The representation
    in this case is:

       $$ LC_i:[LC_i^1 \,\,(S_j), \cdots, LC_i^m \,\, (S_k)] $$

  The first element of the chain $LC_i$ is  the word $ LC_i^1$, and occurs in the sentence  $S_j$ ,  the last  element of the chain $LC_i$ is  the word $ LC_i^m  $, and occurs in the sentence  $S_k$ of the text (where $j < k$).

  LCs   could further serve  as a basis for  Text segmentation and  Text summarization
  (see \cite{book}).  The first paper which used LCs (manually built)  to indicate the structure of a text was that of Morris and Hirst (\cite{Morris and Hirst 1991}), and it relies on the hierarchical structure of Roget's thesaurus to find semantic relations between words.
Since the chains are used to structure the text according to the attentional/intentional theory of Grosz and Sidner theory, (\cite{Grosz and Sidner 1986}), their algorithm divides texts into segments  which form hierarchical structures (each segment is represented by the span of a LC). Some algorithms for linear segmentation (as opposite to hierarchical segmentation)  are given in \cite{Tatar et al. 2008a}, \cite{Tatar Kept}, \cite{Tatar CAI}. In all these algorithms it is applied the following remark of Hearst 1997 \cite{Hearst 1997}: {\it There are certain points at which there may be radical changes in space, time, character configuration, event structure(...). At points where all of these change in a maximal way, an episode boundary is strongly present}. The algorithms are based on different ways of scoring the sentences of a text and then observing the graph of the score function. In this paper we introduce two new scoring functions for sentences (in the next subsection).

Let us remark   that linear segmentation and the (extractive) summarization are two interdependent goals: good segmentation of a text could improve the summarization (\cite{book}).
      Moreover, the rule of extracting
   sentences from the segments is decisive for the  quality of the summary. Some largely applied strategies (rules) are (\cite{Tatar et al. 2008a}):

    ${\bf 1}$. The first  sentence of a segment is selected.

    ${\bf 2}$. For each segment the sentence with a  maximal score  is considered the most important for this segment, and hence it is selected (for example, the minima in the graph of  the below $Score^1$ and $Score^2$ functions represent the sentences candidates for boundaries between segments of a text).

    ${\bf 3}$. From each segment the most informative sentence (the least similar) relative to the previously selected sentences  is picked up.

     Thus, one can say that determining a segmentation of a text and selecting a strategy (1, 2 or 3),  a summary of the  text
can be obtained, as well.

\subsection{Coreference Chains}

  Coreference Chains are  {\it chains of  antecedents-anaphors} of a text. A complete study of Coreference Chains is the textbook \cite{Mitkov 2002}.  A Coreference Chain  contains the occurrences of the entities identified as
antecedents for a given anaphor and also the occurrences of this anaphor.

   The formalization of a Coreference Chain is as follows:

     $$ CR_i:[CR_i^1 \,\,(Token_j), \cdots, CR_i^m \,\, (Token_k)] , (where\, j < k\,) $$

       or
       $$ CR_i:[CR_i^1 \,\,(S_j), \cdots, CR_i^m \,\, (S_k)], (where\, j < k\,) $$

\noindent         depending on the marks (tokens or sentences) picked out.

  In the same way as the Lexical Chains,  Coreference Chains
express the cohesion of a text. The algorithms of segmentation (and summarization) of a text based on  Lexical Chains could be adapted for Coreference Chains. In this paper we refer to both Lexical Chains and   Coreference Chains   by the name of Cohesion Chains.

\subsection{Scoring the sentences by Cohesion Chains}

Cohesion Chains (CCs) defined  as in the above sections could  be used to score the sentences such that when this score is low, cohesion is low, and thus the sentence is a candidate for a boundary between  segments; similarly for a high score (a high cohesion) and the non-boundary feature  of a sentence.  In this paper we propose the following two new functions of score for sentences:



 $$ Score^1(S_i)=  \frac{the\, number\, of \,tokens \,  in \,S_i \,contained \,in\, at\, least\, one \,CC \, }{the\,  number \,of \, tokens\, in \,S_i.}$$

   Let us remark that $ 0\leq Score^1(S_i) \leq 1$. When $ Score^1(S_i)=$   0 (or close to 0), $S_i$ is a candidate for a boundary between  segments because $S_i$ has a low connection with  other sentences. When $ Score^1(S_i)=$   1 (or close to 1), $S_i$ is "very" internal for a segment. So, observing the graph of function  $ Score^1(S_i)=$ we could determine the segments of a text.

  The second proposed scoring function is:

 $$ Score^2(S_i)=  \frac{the\, number\, of \,CCs \,which \,traverse\, S_i} { the\, total\, number\, of\,  CCs\, in\, the\, text}$$

  Again $ 0\leq Score^2(S_i) \leq 1$ and the above remarks remain valid: when $ Score^2(S_i)=$  is 0 (or close to 0), $S_i$ is a candidate for a boundary between  segments because $S_i$ has a low connection with the others sentences. When $ Score^2(S_i)=$  is 1 (or close to 1), $S_i$ is "very" internal for a segment.

    As a final remark, let us observe that the hrebs (quasi-hrebs) could be used exactly in the same way to score the sentences: it is enough to put {\it quasi-hrebs} instead of  {\it CCs} in the definitions for $Score^1(S_i)$ and $Score^2(S_i)$. Thus, hrebs (quasi-hrebs) could serve to segment and/or summarize texts.

    In the same way, the indicators 1-8 used in Denotational analysis with hrebs could be extended to CCs. Let us remark that quasi-hrebs (and thus CCs) are defined in the Data-hrebs format. This is accordingly with the definition of Lexical Chains where
the most important (frequent) lexical relation which is present in a Lexical Chain is the repetition \cite{Stokes et al. 2004}. The more frequently a word is repeated in a Lexical Chain, the more important this Lexical Chain is.
        Obtaining CCs from Data-hrebs (duplicates are not eliminated), we will impose  the  condition to a  kernel CC to have
    at least a given number of elements. In other words, a kernel CC must contain  a size bigger than a minimal one.
 Further, the topicality of a kernel CC, text concentration, CC-diffuseness and mean diffuseness of the text could be defined.

Let us mention that nowhere in the Lexical Chains or Coreference Chains literature these kinds of indicators are introduced up to now.




\section{Example in Romanian}

For  the  Eminescu's poem "Lacul"   we will exemplify  {\bf  hrebs, quasi-hrebs and CCs},  and the relationships between them.  We will begin with  the {\bf Rules} for hreb formation in Romanian language \cite{Popescu et al. 2013}.\\

{\bf Rules of hrebs formation for the Romanian language}\\

The Rules for hrebs are of the form: $"a \in  B"$. Here $a$ is an expression containing a special element called {\it pos} indicator which is written in italic ($pos$ is for $part\, of \, speech$). Particularly, $"a"$ could be formed only from the $pos$ indicator.  $"B"$ is a (name for a) given hreb written with capital letters. More exactly, the Rule  $"a \in B"$ means: $"a"$ (or $pos$ indicator of $"a"$) is an element of the hreb $"B"$ . The connection between $"a"$ and $"B"$ will result from the word used for $pos$ indicator.
      As a word-form could be contained in more then one hreb, in the application of rules it is possible to obtain a result as: $"a \in  B,C,\cdots "$ meaning:  $"a"$ is an element of hreb $"B"$ and hreb $"C"$ and  $\cdots$.
      The rules are valid only for the  $pos$ of $"a"$ being   $noun, \, verb,\, adjective,\, adverb,\, pronoun$.

      {\bf RULES}:

R1. $verb \in   VERB$

R2. personal ending of a $verb$, which could be a  $noun$  or a $pronoun$,  $ \in   NOUN \, or  \,PRONOUN$

R3. synonym of a $verb \in   VERB$

R4. $pronoun$ referring to a $noun$  $\in  NOUN$

R5. $pronoun$ referring to  a Named Entity $\in NAMED\,\, ENTITY$.

R6. synonym of a Named Entity  $\in NAMED\,\, ENTITY$.

R7. non-referring $pronoun \in  PRONOUN$

R8. $noun  \in  NOUN$

R9. synonym of a $noun$ $\in NOUN$



R10. $adjective \in  ADJECTIVE$

R11. synonym of an $adjective \in ADJECTIVE$

R12. $adverb \in  ADVERB$

R13.synonym of an $adverb \in ADVERB$\\


The Rules 1-13 could be summarized as follows: a noun, its synonyms, referring pronouns and personal endings in a verb belong all to the given noun; a Named Entity, its synonyms, referring pronouns and personal endings in a verb belong all to the given Named Entity; a verb in all its forms, its synonyms, belong to the given verb, however, the personal endings belong also to the respective noun; an adjective (adverb) and  its synonyms belong all to the given adjective (adverb).

	We illustrate the rules as applied to the poem  "Lacul". Namely, we will make a denotation of tokens in the poem, then    will extract:
 \begin{itemize}

\item  {\bf A. Hrebs} ( Table 1),

\item {\bf  B. Quasi-hrebs} (Table 2)

\item {\bf C. Cohesion Chains} (Table 3).

\end{itemize}

The tokens numbered are only  nouns, verbs, adjectives, adverbs, and pronouns (in this poem do not  exist Named Entities).\\

    \hspace{4cm}                    {\bf  LACUL} (denotation of tokens)\\

\vspace{0.3cm}

{\bf (S1)} Lacul (1)  codrilor (2)  albastru (3)

Nuferi (4) galbeni (5)  \^ il (6)  \^ incarc\u a (7).

{\bf (S2)}Tres\u arind (8) \^ in cercuri (9)  albe (10)

El (11) cutremur\u a (12) o barc\u a (13).

{\bf (S3)} \c Si eu (14) trec (15)  de-a lung (16)  de maluri (17),

Parc-ascult (18)  \c si parc-a\c stept (19)

Ea (20) din trestii (21)  s\u a r\u asar\u a (22)

\c Si s\u a-mi (23) cad\u a (24) lin (25) pe piept (26).

{\bf (S4)} S\u a s\u arim (27) \^ in luntrea (28) mic\u a (29) ,

\^ Ing\^ ina\c ti (30) de glas (31)  de ape (32),

\c Si s\u a scap (33) din m\^ an\u a (34) c\^ arma (35),

\c Si lope\c tile (36) s\u a-mi (37)  scape (38).

{\bf (S5)} S\u a plutim (39) cuprin\c si (40) de farmec (41)

Sub lumina (42) bl\^ indei (43) lune (44).

{\bf (S6)}V\^ intu-n (45)  trestii (46) lin (47)  fo\c sneasc\u a (48),

Unduioasa (49) ap\u a (50) sune (51)!

{\bf (S7)}Dar nu vine (52)... {\bf (S8)}Singuratic (53)

\^ In zadar (54) suspin (55)  \c si suf\u ar (56)

L\^ ing\u a lacul (57) cel albastru (58)

\^ Inc\u arcat (59) cu flori (60) de nuf\u ar (61).\\

\subsection{From Hrebs to Cohesion Chains}
By the application of the above mentioned rules a total number of 51 hrebs are obtained. From all these, only 12 hrebs presented in Table 1 contain more than one element. In Table 1 the hrebs are constituted as Data-hrebs, where SDH means "Size of Data-hreb" and SSH means "Size of Set-hreb".

 The names of all 51 hrebs are as follows:

A ASCULTA, A A\c STEPTA, A \^ INC\u ARCA, A C\u ADEA, A CUTREMURA, A FO\c SNI,  A P\u AREA, ,  A PLUTI,  A R\u AS\u ARI, , A S\u ARI, A SC\u APA, A SUFERI, A SUNA, A SUSPINA, A TRECE, A TRES\u ARI, A VENI, ALB, ALBASTRU, AP\u A, BARC\u A, BL\^ AND\u A, C\^ ARM\u A, CERC, CODRU, CUPRINS, EU, EA, FARMEC, FLOARE, GALBEN, GLAS,  \^ INC\u ARCAT, \^ ING\^ ANAT,   LAC, LIN, LOPAT\u A, LUMIN\u A, LUN\u A, LUNG, MAL, M\^ AN\u A, MIC, NOI, NUF\u AR, PIEPT, SINGURATIC, TRESTIE, UNDUIOAS\u A, V\^ ANT, ZADAR.

 \begin{table}

{\small
\begin{center}

\begin{tabular}{*{4}{|c}|}
 \hline
\multicolumn{1}{|c|}{Hreb} &
\multicolumn{1}{|c|}{Elements of Data-hreb} & \multicolumn{1}{|c|}{SDH} &
\multicolumn{1}{|c|}{SSH} \\

\hline EU &  {\footnotesize (eu 14, {\bf trec  15, -ascult 18, -a\c stept 19,
 scap  33}, -mi 23,}\\

\hline (EU cont) & {\footnotesize -mi 37 , {\bf suspin  55, suf\u ar 56}})
 & $9$  & $ 8$  \\

\hline LAC & (lacul\,1, il\, 6, {\bf tres\u arind\, 8}, el \, 11, {\bf cutremur\u a \, 12}, lacul 57)  & $6$ & $5$  \\

\hline EA & (ea\, 20,  {\bf r\u asar\u a \, 22, cad\u a  \,24,  vine \, 52}) & $4$ & $
4$ \\

\hline NUF\u AR & (nuferi \, 4, {\bf incarc\u a} \, 7, nuf\u ar \, 61) & $3$ & $ 2$  \\

\hline AP\u A & (ape \, 32, ap\u a \, 50, {\bf sune} \,51 )  & $3$  & $ 2$  \\

\hline NOI & ({\bf s\u arim \, 27, plutim \, 39})   & $2$ & $ 2$  \\

\hline BARC\u A & (barca \, 13, luntrea \, 28)   & $2$  & $2$ \\

\hline TRESTIE & (trestii \, 21, trestii\, 46) & $2$ & $1$  \\

\hline ALBASTRU & (albastru \, 3, albastru \, 58)  & $2$ &  $1$  \\

\hline A \,P\u AREA & (parc- \, 18, parc-\, 19) & $2$ & $1$  \\

 \hline LIN & (lin \, 25, lin \, 47) & $ 2$ &   $1$  \\

\hline A\, SC\u APA & (scap\, 33, scape\, 38) & $2$ &   $1$  \\

\hline

\end{tabular}
\end{center}
}
\caption{{\bf A.} The hrebs  with size bigger than 1 extracted from the poem {\bf Lacul}}

\label{t1}

\end{table}


  From the set of Rules R1-R13, the  Rule R2 makes the difference when the quasi-hrebs are calculated. This rule is reproduced here:\\

  R2. personal ending of a $verb$, which could be a  $noun$ or a $pronoun$,  $ \in   NOUN \, or \,PRONOUN$\\

  In Table 1 are bold marked all the verbs which are contained in a NOUN or PRONOUN
   hreb due to the Rule R2. All these verbs are not present in Table 2, the
   table of quasi-hrebs. As a remark, the hreb "NOI" is not a quasi-hreb,
   because both elements   ( s\u arim \, 27, plutim \, 39) are obtained by Rule R2.

\begin{table}

{\small
\begin{center}

\begin{tabular}{*{4}{|c}|}
 \hline
\multicolumn{1}{|c|}{Quasi-hreb} &
\multicolumn{1}{|c|}{Elements of Data-hreb} & \multicolumn{1}{|c|}{SDH} &
\multicolumn{1}{|c|}{SSH} \\

\hline EU &  (eu 14, , -mi 23,
-mi 37  )
 & $3$  & $ 2$  \\

\hline LAC & (lacul\,1, il\, 6, , el \, 11, , lacul \,57)  & $4$ & $3$  \\

\hline EA & (ea\, 20) & $1$ & $
1$ \\

\hline NUF\u AR & (nuferi \, 4,   nuf\u ar \, 61) & $2$ & $ 1$  \\

\hline AP\u A & (ape \, 32, ap\u a \, 50)  & $2$  & $ 1$  \\

\hline BARC\u A & (barca \, 13, luntrea \, 28)   & $2$  & $2$ \\

\hline TRESTIE & (trestii \, 21, trestii\, 46) & $2$ & $1$  \\

\hline ALBASTRU & (albastru \, 3, albastru \, 58)  & $2$ &  $1$  \\

\hline A \,P\u AREA & (parc- \, 18, parc-\, 19) & $2$ & $1$  \\

 \hline LIN & (lin \, 25, lin \, 47) & $ 2$ &   $1$  \\

\hline A\, SC\u APA & (scap\, 33, scape\, 38) & $2$ &   $1$  \\

\hline

\end{tabular}
\end{center}
}
\caption{{\bf B.} The quasi-hrebs extracted from the poem {\bf Lacul}}

\label{t1}

\end{table}


   Let us remember that Lexical Chains  are sequences of words which are in a lexical cohesion relation (synonymy, repetition, hypernymy,  hyponymy, etc) with each other.  Coreference Chains are  chains of  antecedents-anaphors of a text.
   Examining Table 2 of quasi-hrebs, we observe that:
   the quasi-hreb EU corresponds to a Coreference Chain (eu 14,  -mi 23,
-mi 37), the quasi-hreb LAC  to a Coreference Chain (lacul\,1, il\, 6, \,  el \, 11,  \, lacul \,57). The quasi-hreb EA is not a chain (it has only one element). The rest of quasi-hrebs represents Lexical Chains:  (nuferi \, 4,   nuf\u ar \, 61),
(ape \, 32, ap\u a \, 50),
 (barca \, 13, luntrea \, 28),
 (trestii \, 21, trestii\, 46),
 (albastru \, 3, albastru \, 58),
 (parc- \, 18, parc-\, 19),
 (lin \, 25, lin \, 47),
 (scap\, 33, scape\, 38).
Table 3 contains the  Cohesion Chains denoted as we will use further. We obtained CCs from the Data-hrebs, and the length of a Cohesion Chain is given by the SDH column, because the duplicates are not eliminated (as in SSH column). \\

\begin{table}

{\small
\begin{center}

\begin{tabular}{*{3}{|c}|}
 \hline
\multicolumn{1}{|c|}{Denotation of CC} &
\multicolumn{1}{|c|}{Elements of CC} & \multicolumn{1}{|c|}{Length of CC} \\

\hline CC1 &  (eu 14,  -mi 23,
-mi 37  )
 & $3$    \\

\hline CC2 & (lacul\,1, il\, 6, \, el \, 11, , lacul \,57)  & $4$   \\

\hline CC3 & (nuferi \, 4, \,  nuf\u ar \, 61) & $2$   \\

\hline CC4 & (ape \, 32,\, ap\u a \, 50)  & $2$   \\

\hline CC5 & (barca \, 13,\, luntrea \, 28)   & $2$   \\

\hline CC6 & (trestii \, 21,\, trestii\, 46) & $2$   \\

\hline CC7 & (albastru \, 3, albastru \, 58)  & $2$   \\

\hline CC8 & (parc- \, 18, parc-\, 19) & $2$   \\

 \hline CC9 & (lin \, 25, lin \, 47) & $ 2$   \\

\hline CC10& (scap\, 33, scape\, 38) & $2$   \\

\hline

\end{tabular}
\end{center}
}
\caption{{\bf C.} Cohesion Chains  extracted from the poem {\bf Lacul}}

\label{t1}

\end{table}

Calculating the scores $Score^1$ for each sentence are obtained the following results:\\

$Score^1(S_1)= 4/7=0.57 $

$Score^1(S_2)= 2/6=0.33$

$Score^1(S_3)= 6/13=0.46 $

$Score^1(S_4)= 5/12=0.42 $

$Score^1(S_5)= 0/6=0. $

$Score^1(S_6)= 3/7=0.43 $

$Score^1(S_7)= 0/1=0. $

$Score^1(S_8)= 3/9=0.33 $\\

Taking as segment boundaries  the sentences with minimal score, the text is divided in 4  segments: $Seg1=[S_1,S_2];Seg2=[S_3,S_5];Seg3=[S_6,S_7];Seg4=[S_8]$ or 3 segments:
$Seg1=[S_1,S_2];Seg2=[S_3,S_5];Seg3=[S_6,S_8]$ if mono-sentence segments are not permitted.  \\

Scoring with $Score^2$ formula, the results are as following:\\

$Score^2(S_1)= 3/10=0.30 $

$Score^2(S_2)= 4/10=0.40 $

$Score^2(S_3)= 8/10=0.80 $

$Score^2(S_4)= 9/10=0.90 $

$Score^2(S_5)= 6/10=0.60 $

$Score^2(S_6)= 6/10=0.60 $

$Score^2(S_7)= 3/10=0.30 $

$Score^2(S_8)= 3/10=0.30 $\\

 The text has only one segment $[S_1,S_8]$, with the most "internal" sentence $S_4$.
A summary of the poem using  $Score^1$ is formed by the sentences: $S_1, S_3, S_6$ and using
$Score^2$, by the sentence $S_1$. In both cases the rule one  (Section 3.1) has been applied.

\subsection{Indicators of Cohesion Chains}

 Let us suggest how  the indicators in Section 2.1 could be defined for  the Cohesion Chains CC1 to CC10.

 \begin{itemize}

 \item Kernel CCs : Considering the minimal size of a kernel CC being 2, all CCs are in $Kernel$. Considering the minimal size of a kernel CC being 3, only CC1 and CC2 are in $Kernel$. The last supposition is more realistic, since a CC has always at least 2 elements;

\item The size of the $Kernel$ is 2, in the last above case;

\item  Topicality of the kernel CC denoted by CC1 is $3/2=1.5$ and topicality of CC2 is $4/2=2$;

\item Kernel concentration is $KC=2/10=0.2$;

\item $p_1=3/61; p_2=4/61; p_i=2/61,\, i=3\, to\, 10. $
          Text concentration is $TC=0.0151$ and Relative Text concentration is $TC_{Rel}=1.2830$;

\item  Diffuseness for each CC is as follows:

        $D_{CC1}=(37-14)/3=7.66;D_{CC2}=(57-1)/4=14; D_{CC3}=(61-4)/2=28.5; D_{CC4}=(50-32)/2=9; D_{CC5}= (28-13)/2=7.5;D_{CC6}=(46-21)/2=12.5;D_{CC7}=(58-3)/2=27.5; D_{CC8}=(19-18)/2=0.5; D_{CC9}=(47-25)/2=11; D_{CC10}=(38-33)/2=2.5$

\item Mean diffuseness of the text is $D_{Text}=10.75$;

\item Text compactness is $ C=(1-10/61)/(1-1/61)= 0.8505.$

\end{itemize}

 The above  indicators could make  differences between CCs, such that some of them are kernel CCs, or have a higher topicality and/or diffuseness.

\section{Conclusions and Further work}

   Lexical Chains and Coreference Chains (CCs) are intensively studied, but few indicators are standard for them. The indicators inspired from the hrebs must be studied and adopted for CCs.  These indicators, in the context of some applications using CCs, could  become instruments for the evaluation  of these applications and for improving them.
   For example, there is a large debate about how to  select CCs to construct the summaries of a text: selecting long or short CCs is  one of the  questions.  Using only kernel CCs, or
   kernel CCs with a high topicality and /or  high diffuseness could be a solution.

    As a general remark, Quantitative Linguistics and Computational Linguistics
     are considered two distinct fields with their own journals, techniques and specialists.
           It is important  to identify
           those parts they have in common,
           and to try to extract the advantage from this commonality. This paper is a step toward this desirable aim.

\label{finalpage}
\end{document}